\documentclass[pdflatex,sn-mathphys-num]{sn-jnl}

\usepackage{graphicx}%
\usepackage{multirow}%
\usepackage{amsthm}%
\usepackage{mathrsfs}%
\usepackage{wrapfig}
\usepackage{amsmath}
\usepackage{amssymb}
\usepackage{mathtools} 
\usepackage[title]{appendix}%
\usepackage[usenames,dvipsnames]{xcolor}  
\usepackage{textcomp}%
\usepackage{manyfoot}%
\usepackage{booktabs}%
\usepackage[ruled,vlined]{algorithm2e}%
\usepackage{listings}%
\usepackage{pgf}

\usepackage{pifont}
\definecolor{darkgreen}{RGB}{100,200,100}
\usepackage{booktabs} 
\usepackage{tabularx}

\theoremstyle{thmstyleone}%
\theoremstyle{thmstyletwo}%

\theoremstyle{thmstylethree}%

\newcommand\syllogism[3][]{%
  \begin{center}
  \def\tmp{#1}%
  \ifx\tmp\empty\else(#1)\quad\fi
  \begin{tabular}{@{}l@{}}
    #2\\\hline#3\quad$\therefore$
  \end{tabular}
  \end{center}
}

\begin{document}

\title[Article Title]{Data-driven Machine Learning Cannot Reach Symbolic-level Logical Reasoning -- The Limit of the Scaling Law}


\author[1,2]{\fnm{Tiansi} \sur{Dong}}\email{tdong@turing.ac.uk}\email{td540@cam.ac.uk}

\author[2]{\fnm{Mateja} \sur{Jamnik}}\email{mj201@cam.ac.uk} 

\author[2]{\fnm{Pietro} \sur{Liò}}\email{pl219@cam.ac.uk} 

\affil[1]{ \orgname{The Alan Turing Institute}, \orgaddress{\street{96 Euston Road}, \city{London}, \postcode{NW1 2DB}, \country{UK}}}

\affil[2]{\orgdiv{Department of Computer Science and Technology}, \orgname{University of Cambridge}, \orgaddress{\street{15 JJ Thomson Ave.}, \city{Cambridge}, \postcode{CB3 0FD}, \country{UK}}}


\abstract{
By promoting vectors to spheres and enabling explicit model construction, neural networks can perform symbolic-level syllogistic reasoning without training data. We identify two fundamental limitations that prevent conventional data-driven machine learning systems from achieving this capability: training data generated by the combination table cannot distinguish all 24 valid syllogism types, and end-to-end premise-to-conclusion mapping creates contradictory targets within neural components. Experiments with two representative conventional systems—GPT-5 using linguistic inputs and Euler Net using visual inputs—support this analysis. ChatGPT GPT-5 may reach 100\% accuracy in syllogistic reasoning, but with hallucinations. Because the learning process terminates upon reaching 100\% accuracy, the system cannot progress beyond empirical accuracy to symbolic-level reasoning. Random test data reduced Euler Net’s accuracy to 56\%. Repeatedly expanding the training set increased its accuracy to 97\%, with perfect performance on 8 syllogism types. However, because unintended inputs cannot be exhaustively covered, even 100\% test accuracy does not imply symbolic-level reasoning. Since syllogistic reasoning underpins logical reasoning and human rationality, these results suggest that increasing data and training time alone cannot ensure symbolic-level logical reasoning.
}

\maketitle

\section{Introduction}

The historical success of neural networks, particularly LLMs, has been witnessed in various applications, such as human-like communication~\citep{chatgpt_nature2023}, playing games~\citep{alphaGo2017,alphaGo2020}, predicting gene structures~\citep{AlphaFold3}, and solving mathematical tasks~\citep{Davies21,alphaproof2024}. By increasing the amount of training data and training time~\citep{scalinglaw2020,scalinglaw24}, data-driven machine learning systems may steadily enhance their reasoning capabilities. However, their reasoning abilities are still limited, even for simple logical reasoning~\citep{chatgpt_nature2023}. The central notion of logical reasoning, from the origin of logic research in history till now, is the notion of ``following from”, or more formally, ``logical consequence from the premises” – what can we know from the premises? ~\citep{historyLogic17}. This suggests that neural networks capable of symbolic-level logical reasoning should not rely on training data, as demonstrated in syllogistic reasoning by Sphere Neural Networks~\citep{djl2024sphere,djl2025,djl2026monkey}, a specialised class of RNNs. This result is perhaps unsurprising, given that RNNs are Turing complete~\citep{turingcomp23,transformerREP24}. However, this raises the question of whether data-driven machine learning systems can achieve—or approach arbitrarily closely—the same level of performance by increasing the volume of training data and the duration of training. 

Syllogistic reasoning can take either linguistic or visual form. Here, we show that data-driven machine learning systems cannot achieve symbolic-level syllogistic reasoning in either modality. Because syllogistic reasoning underpins logical reasoning, we conclude that they cannot attain symbolic-level logical reasoning.

The paper is structured as follows: We first introduce syllogistic reasoning in linguistic and visual forms; Then, we survey data-driven machine learning systems for syllogistic reasoning, focusing on LLMs and Euler Net, as representative systems in either modality. We identify two methodological limitations in both modalities that prevent them from achieving symbolic-level syllogistic reasoning, and conduct a series of experiments confirming that increasing the amount of training data alone is unlikely to provide a path toward symbolic-level reasoning.

\begin{figure*}[t]
  \centering
\includegraphics[width=1\textwidth]{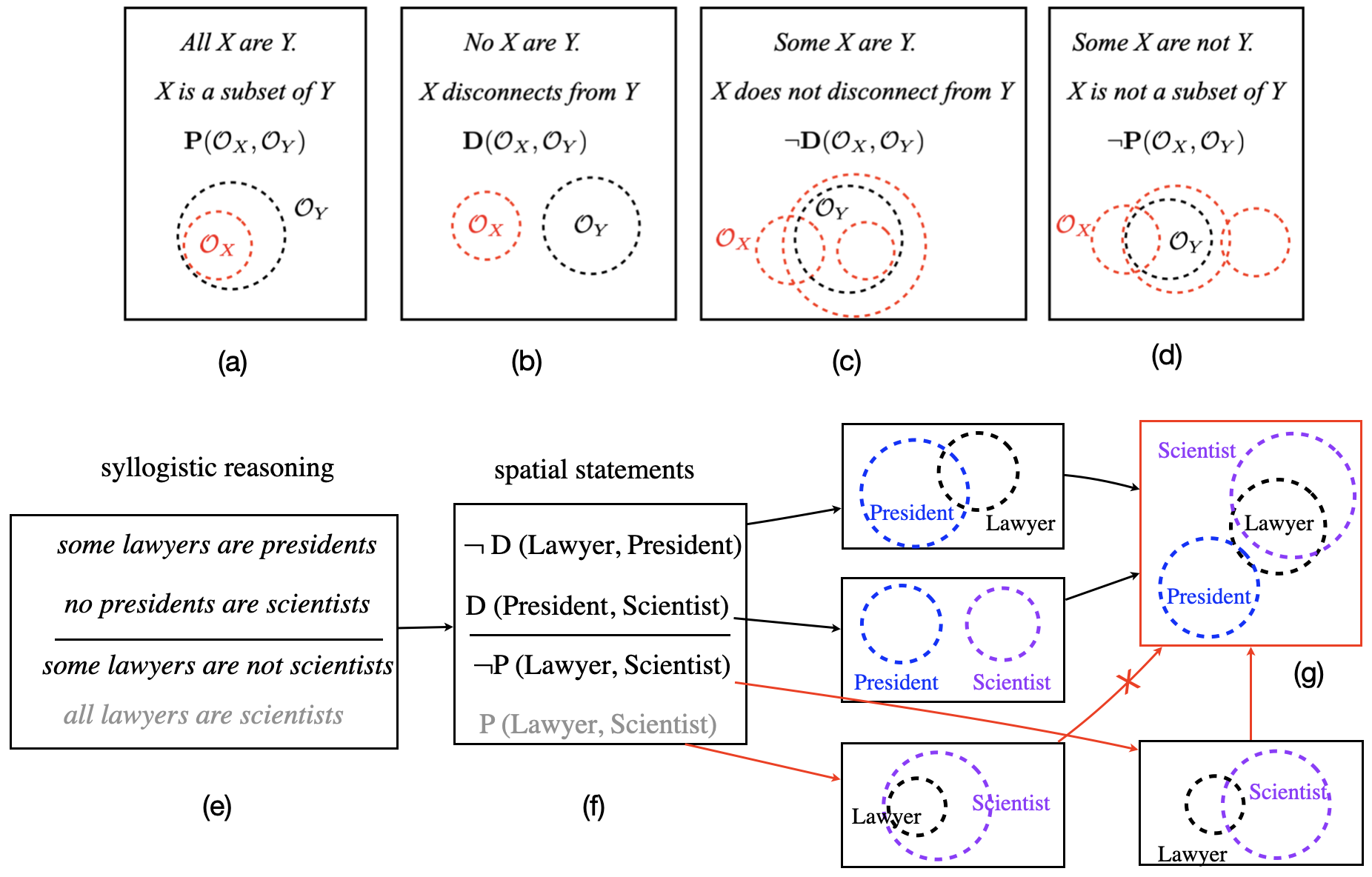}
  \caption{(a-d) Four syllogistic relations and their spatial relations; (e) from the two premises, the logical conclusion is {\em some lawyers are scientists}, its negation is {\em all lawyers are scientists}; (f) spatial statements of the syllogistic statements; (g) no sphere configuration satisfies the premises and the conclusion $\mathbf{P}$(lawyers, scientists); there is a sphere configuration that satisfies the premises and the conclusion $\neg\mathbf{P}$(lawyers, scientists)}
\label{syllogism_intro}
\end{figure*}
\section{Linguistic and Visual Syllogistic Reasoning}
\label{sec_syl}
Syllogistic reasoning, developed by Aristotle more than 2,000 years ago, marked the beginning of formal logical reasoning, which later evolved through propositional logic to first-order logic~\citep{historyLogic17}. A syllogistic reasoning is a deductive argument consisting of two premises and a conclusion involving only three terms. The relations among these terms are restricted to four possible forms: (1) all $X$ are $Y$; (2) some $X$ are $Y$; (3) no $X$ are $Y$; (4) some $X$ are not $Y$. The four relations can be interpreted in terms of set relations and visually represented in Euler diagrams, as shown in Figure~\ref{syllogism_intro}(a-d). For example, {\em some X are Y} can be interpreted as the relation ``set X ($\mathcal{O}_X$) intersects with set Y ($\mathcal{O}_Y$)'', which corresponds to three possible diagrammatic relations: (1) $\mathcal{O}_X$  partially overlaps with $\mathcal{O}_Y$, (2) $\mathcal{O}_X$ contains $\mathcal{O}_Y$, (3) $\mathcal{O}_Y$ contains $\mathcal{O}_X$. We can merge the three possible relations into one relation: $\mathcal{O}_X$ does not disconnect from $\mathcal{O}_Y$, $\neg\mathbf{D}(\mathcal{O}_X, \mathcal{O}_Y)$, as shown in Figure~\ref{syllogism_intro}(c).  
Syllogistic relations can be elegantly defined using the part-of relation $\mathbf{P}$~\citep{Smith96}, establishing one-to-one correspondence between syllogistic and diagrammatic relations as follows.  
\begin{itemize}
    \item ``all $X$ are $Y$'' $\Leftrightarrow$ ``$\mathcal{O}_X$ is part of $\mathcal{O}_Y$'', $\mathbf{P}(\mathcal{O}_X, \mathcal{O}_Y)$;
    \item ``some $X$ are $Y$'' $\Leftrightarrow$ ``$\mathcal{O}_X$ does not disconnect from $\mathcal{O}_Y$'', $\neg\mathbf{D}(\mathcal{O}_X, \mathcal{O}_Y)$;
    \item ``no $X$ are $Y$'' $\Leftrightarrow$ ``$\mathcal{O}_X$ disconnects from $\mathcal{O}_Y$'', $\mathbf{D}(\mathcal{O}_X, \mathcal{O}_Y)$;
    \item ``some $X$ are not $Y$'' $\Leftrightarrow$ ``$\mathcal{O}_X$ is not part of $\mathcal{O}_Y$'', $\neg\mathbf{P}(\mathcal{O}_X, \mathcal{O}_Y)$.
\end{itemize}
Formally, we define $\mathcal{O}_X$ disconnecting from $ \mathcal{O}_Y$ as that ``there is no $\mathcal{O}_Z$ that is part of $\mathcal{O}_X$ and $\mathcal{O}_Y$''.  
$$\mathbf{D}(\mathcal{O}_X, \mathcal{O}_Y)\triangleq \nexists\mathcal{O}_Z [\mathbf{P}(\mathcal{O}_Z, \mathcal{O}_X)\land\mathbf{P}(\mathcal{O}_Z, \mathcal{O}_Y)]$$
If we allow two terms in premises to change positions and fix the order of terms in the conclusion statement, there will be 256 different forms of Aristotelian syllogistic reasoning, among which 24 types (listed in the Appendix) are {\em valid}~\citep{laird2012}.

A syllogistic reasoning can be {\em satisfiable, unsatisfiable, valid}, or {\em invalid}. 
Being {\em satisfiable} means there is a case in which both the premises and the conclusion are true. Being {\em valid} means the conclusion is true in every case its premises are true \citep{jeffrey81}. 
For a {\em valid} reasoning, the negation of its conclusion is {\em unsatisfiable}; for an {\em invalid} reasoning, the negation of its conclusion is {\em satisfiable}. Diagrammatically, 
syllogistic reasoning is {\em satisfiable}, if and only if we can construct an Euler diagram, e.g., three circles arranged to satisfy the spatial relations specified by the premises and conclusion; otherwise, this reasoning will be {\em unsatisfiable}.  For example, given the premises {\em some lawyers are presidents} and {\em no presidents are scientists}, the conclusion is {\em some lawyers are not scientists}, while its negation is {\em all lawyers are scientists}, as shown in Figure~\ref{syllogism_intro}(e). In Figure~\ref{syllogism_intro}(g), we successfully constructed an Euler diagram of the premises and the conclusion {\em some lawyers are not scientists}. However, no Euler diagram can simultaneously satisfy the premises and the conclusion {\em all lawyers are scientists}. Thus, this conclusion is {\em unsatisfiable}, and its negation is therefore {\em valid}.

A reasoning network achieves the symbolic level of syllogistic reasoning if it can correctly determine any {\em valid} syllogistic inference and construct counter-examples to {\em invalid} ones. This criterion also applies to data-driven machine learning systems over out-of-distribution data.

\section{Data-driven Machine Learning Systems for Syllogistic Reasoning}
\label{sec_en}

As a basic logical deduction, syllogistic reasoning is straightforward for symbolic methods~\citep{VukmirovicBCS19,BentkampBTV21}. However, developing neural syllogistic models is extremely challenging, to the point that it was regarded as utopian a decade ago~\citep{laird2012}. 
Considering the scaling law~\citep{scalinglaw2020,scalinglaw24} and the Turing Completeness of RNNs~\citep{SIEGELMANN1995,turingcomp23}, our research question is: Can data-driven machine learning attain—or converge arbitrarily closely to—the symbolic level performance as the amount of training data approaches infinity?
The significance of this question lies in the implication of a negative answer: data-driven machine learning cannot attain the rigour of symbolic logical reasoning, because syllogistic reasoning underpins more complex forms of logical reasoning.
 
\subsection{LLMs for syllogistic reasoning}
Data-driven machine learning systems may learn syllogistic reasoning from either linguistic or visual inputs. Here, we focus on LLMs, given their unprecedented success, widespread applications, and use of vast amounts of training data, as representative systems that process linguistic inputs. Several studies have explored LLMs' syllogistic reasoning performance. \cite{Eisape2024} evaluated PaLM 2 family LLMs~\citep{palm22023} and Llama 2 family LLMs~\citep{llama22023}, showing that PaLM 2-Small achieved the best accuracy about 75\%, better than PaLM 2-Large, which does not strictly follow the Scaling Law. 
\cite{syllogism24} evaluated PaLM 2 LLMs and GPT-3.5~\citep{openAI23}, concluding that LLMs may achieve above-chance performances in familiar situations but exhibit numerous imperfections in abstract reasoning, including syllogism. \cite{syllobio2025} examined Mistral LLMs~\citep{mistral7b,mistral23}, Gemma LLMs~\citep{gemma2024}, Llama-3 LLMs~\citep{llama3}), and BioMistral LLMs~\citep{biomistral2024}, with conclusions that zero-shot
LLMs achieved an average accuracy between 70\% on {\em generalised modus ponens} and
23\% on {\em disjunctive syllogism}, and both zero- and few-shot LLMs are sensitive to surface-level lexical variations. Thus, they are far from achieving the reliability required for high-stakes biomedical applications, let alone attaining the rigour of symbolic-level reasoning. \cite{djl2025} evaluated GPT-3.5-turbo and GPT-4 in determining the validity of all types of classic syllogistic reasoning, showing that \mbox{ChatGPT} (GPT-3.5-turbo) reached the best performance (correct decision and explanation) of $46.9\%$ using syllogistic statements with simple symbols, and \mbox{ChatGPT} (GPT-4o) reached the best performance of $82.4\%$ with long random symbols. 
\begin{figure}[t]
  \centering
\includegraphics[width=1\textwidth]{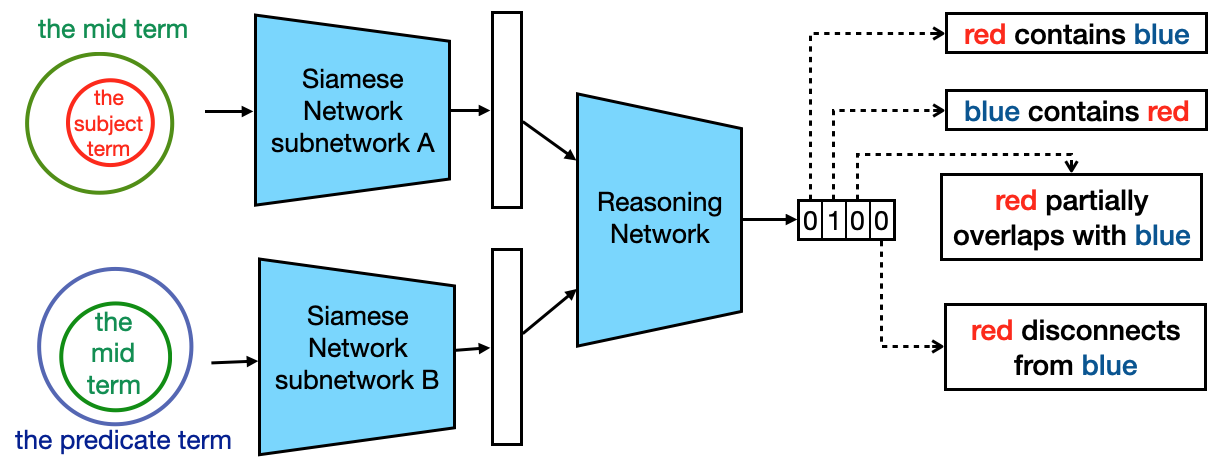}
  \caption{An overview of the Euler Net for syllogistic reasoning. Inputs are two simple images, each consisting of two circles; the output is a vector representing the conclusion.}
  \label{eulernet}
\end{figure}
\begin{figure}
\centering 
\includegraphics[width=1\textwidth]{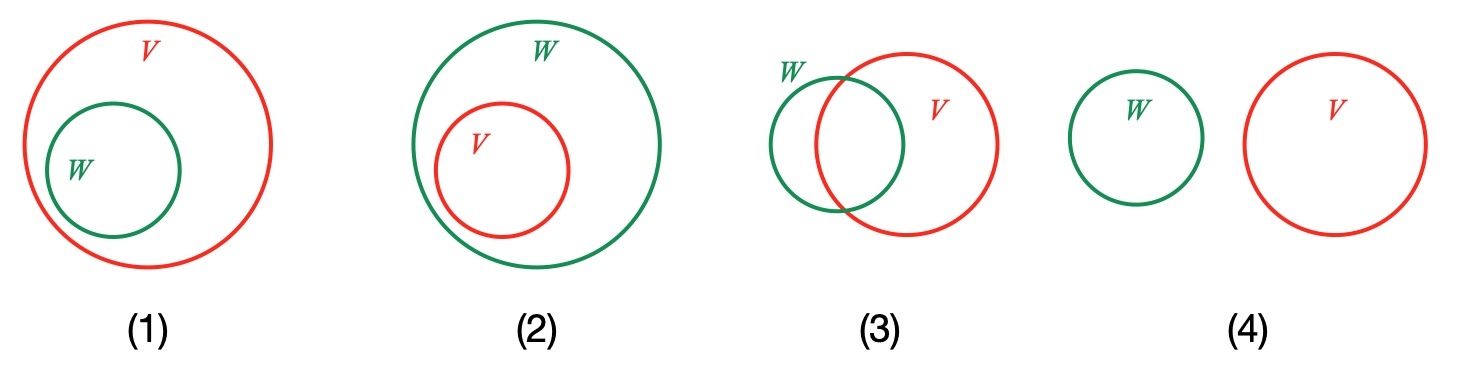}
  \caption{Euler diagrams representing 4 possible relationships between non-empty sets $W$ and $V$.}  \label{euler_diagram}
\end{figure}

\begin{figure*}[t]
  \centering 
\includegraphics[width=1\textwidth]{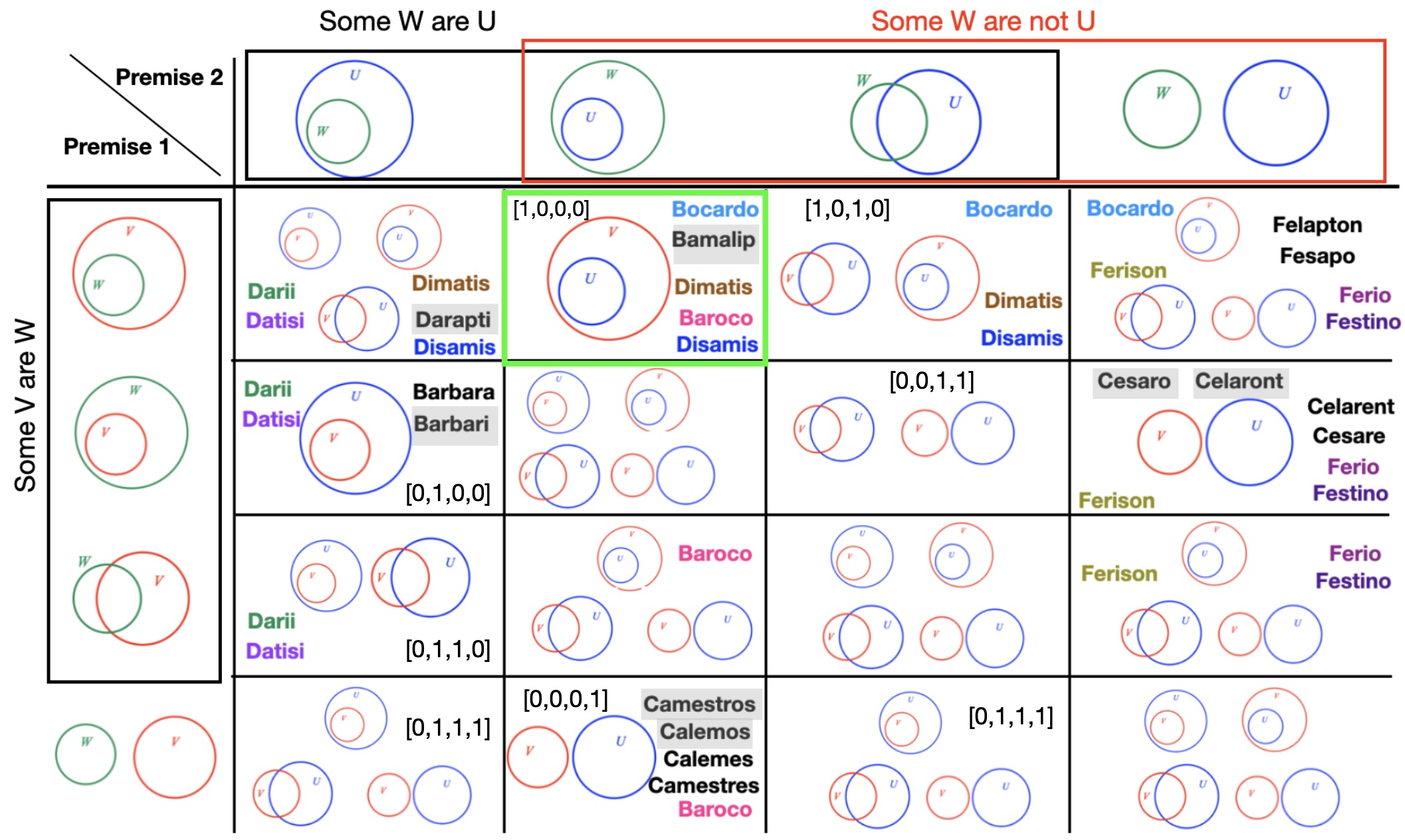}
\caption{Training data for Euler Net is generated by this combination table. Each column and each row are images representing premises. The intersection cell lists all possible conclusions. Euler Net learns syllogistic reasoning by remembering the combinations of Euler diagrams. The combination table establishes associations between inputs (premises) and outputs (conclusions). Premises of {\em Some ... are (not) ...} occupy three columns or rows. The cell with the green boundary hosts 5 valid types of syllogistic reasoning.}
  \label{syl}
\end{figure*}
\subsection{Euler Net for syllogistic reasoning}
A less popular machine learning approach, Euler Net~\citep{WangJL18,WangJL20}, uses image inputs for syllogistic reasoning and achieves almost 100\% accuracy, as illustrated in Figure~\ref{eulernet}. Euler Net is a supervised convolutional deep learning network that solves
syllogistic reasoning by using Euler diagrams, where syllogistic relations are interpreted as basic set relations~\cite{hammer98}, e.g. $W$ is a subset of $V$ ($W\subset V$), as shown in Figure~\ref{euler_diagram}(a). 

The inputs of Euler Net are two Euler diagrams, each
representing one premise; the output is a vector presenting the syllogistic 
conclusion, following the combination table, as shown in Figure~\ref{syl}. For example, let two premises be (1) ``all blue are green''; (2) ``all green are red'', the conclusion will be ``all blue are red'', that is, the red circle contains the blue circle, represented as $[1, 0, 0, 0]$ (each element represents a syllogistic relation, as illustrated in Figure~\ref{eulernet}). With 80000 training patterns and 8000 validation patterns,  Euler Net reached 99.8\% accuracy in 8000 testing data.

\begin{figure*}[t] 
  \centering  
\includegraphics[width=1\textwidth]{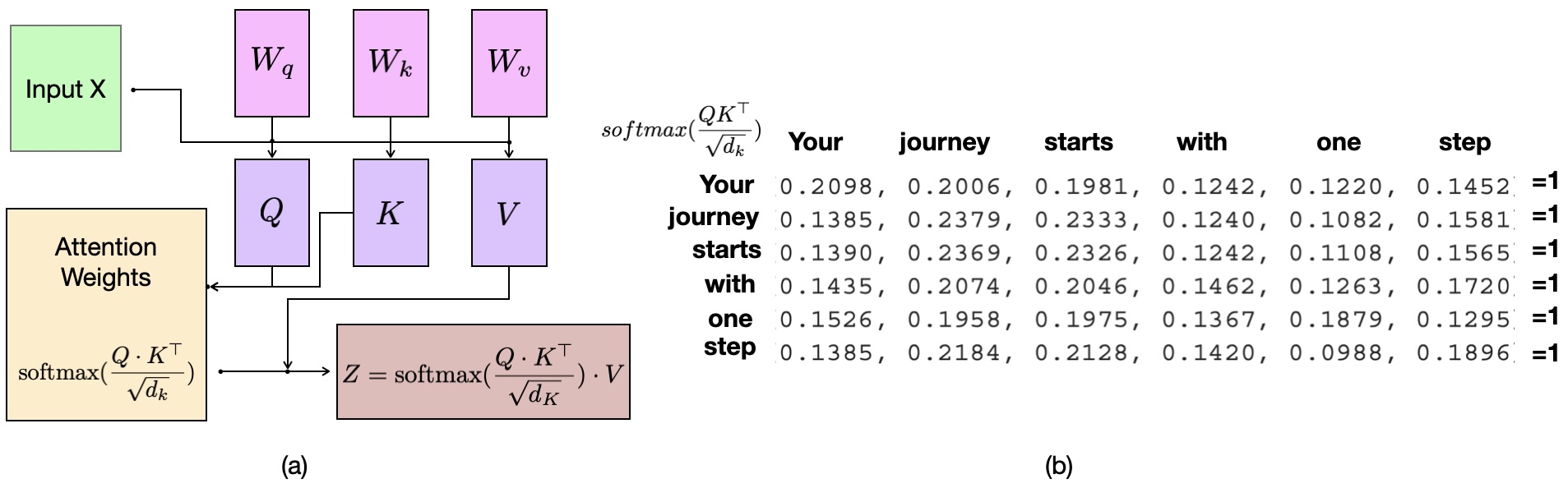}
  \caption{(a) The self-attention mechanism of transformers. $Q$, $K$, and $V$ structures a combination table of words in a sentence; (b) The self-attention values among words in the sentence ``Your journey starts with one step"~\cite{build-llms24}.}
  \label{en}
\end{figure*} 

\section{Methodological Limitations of Data-driven Machine Learning for Logical Reasoning}
\label{method}

EulerNet’s nearly 100\% accuracy may suggest that the remaining 0.2\% gap could be easily closed by adding slightly more training data. 
Contrary to this intuition, we show that no training dataset can expose data-driven systems to every type of valid syllogistic inference and that end-to-end learning introduces inconsistent subtask targets. Together, these limitations prevent such systems from attaining symbolic-level logical reasoning.

\begin{figure}[t]
  \centering
\includegraphics[width=1\textwidth]{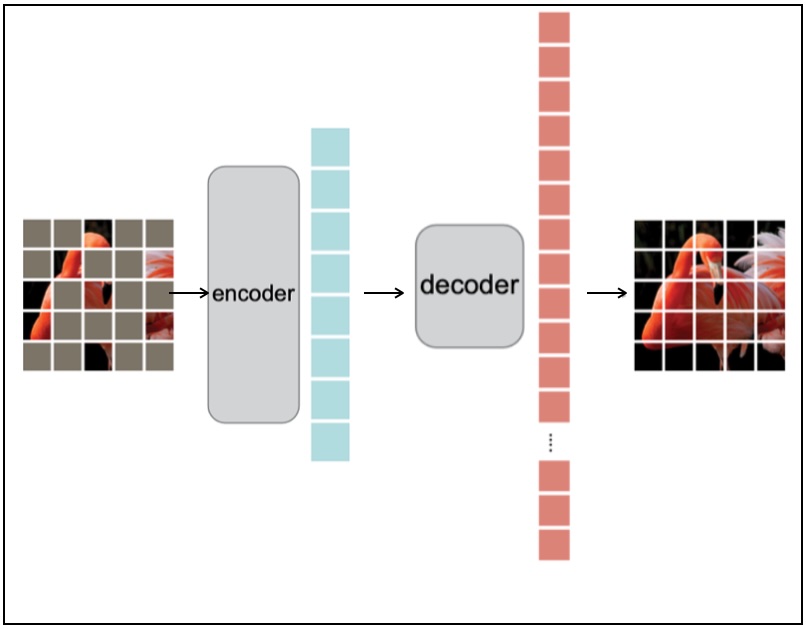}
  \caption{The MAE architecture can reconstruct the whole image from its part, even when the part is only 25\% of the original image~\citep{MAE2022}.}
  \label{mae22}
\end{figure}
\begin{figure}[h]  
\centering
\includegraphics[width=1\textwidth]{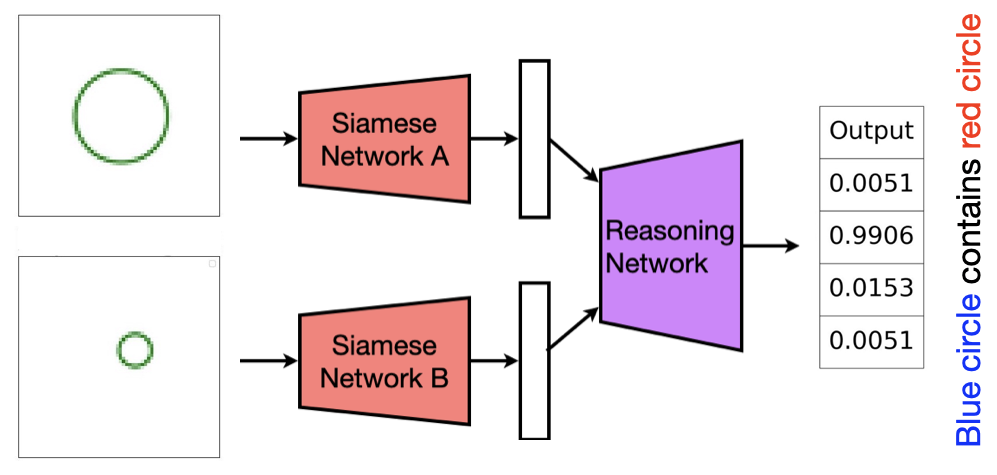}
  \caption{Euler Net may inject blue and red circles into the inputs and predict ``{\color{blue}{blue}} contains {\color{red}{red}}".} 
  \label{br}
\end{figure}

\subsection{Training data cannot distinguish valid syllogisms}
\label{tb}


For data-driven machine learning systems with visual inputs, the syllogistic relations are represented as set-theoretic relations, which pose representation difficulties. The syllogistic statement {\em some $W$ are $V$} has three possible set-theoretic relations: $W\subset V$, $V\subset W$, $W\cap V\neq \emptyset$; and the syllogistic statement {\em some $W$ are not $V$} has another three possible set-theoretic relations: $W\cap V\neq \emptyset$, $V\subset W$, $W\cap V= \emptyset$. This results in the fact that the combination table cannot distinguish each valid type of syllogistic reasoning -- the two syllogistic relations {\em some ... are ...} and {\em some ... are not ...} occupy three rows, columns, and cells, as shown in Figure~\ref{syl}. Though Euler Net demonstrates close to 100\% accuracy on benchmark datasets, its performance is limited to the dataset, which does not distinguish each valid syllogistic reasoning.  

 
For data-driven machine learning systems with linguistic inputs, syllogistic reasoning is learned through learning the correlation between the premises and the conclusion. The self-attention mechanism, used in Transformers~\cite{Vaswani17} and LLMs~\cite{build-llms24}, is highly effective at capturing long-range dependencies. 
It can be understood as learning a combination table (a relation matrix) among the words in a sentence. Each entry in the matrix quantifies the probability that a row word attends to, or is associated with, a column word. The softmax operation converts attention scores into a probability distribution whose values sum to one, as illustrated in Figure~\ref{en}(a,b). By learning statistical regularities from large-scale text corpora, LLMs generate text by selecting words or tokens with the highest probability. Given {\em all A are B. all B are C. Therefore, \_\_ A are C}, a well-trained LLM will complete {\em all A are C}, or {\em some A are C}, or both. Both are correct: the former is the well-known BARBARA type of syllogistic reasoning, while the latter is the BARBARI type, a valid type with the same premises and a weaker conclusion. Because softmax distributes a total probability of one across all candidate words, self-attention alone cannot assign both types the probability of 100\%. It therefore cannot attain symbolic-level syllogistic reasoning.




 
\subsection{End-to-end learning creates conflicting targets}
\label{conflictfeature} 

Training data for data-driven machine learning systems establishes an end-to-end association between premises and conclusion. This end-to-end mapping can be decomposed into subprocesses of latent semantic encoding and decoding. For visual input systems, latent semantic encoding is a neural model for object recognition. A well-trained data-driven machine learning system can recognise an object from its parts, which is a desired feature in Computer Vision~\citep{MAE2022}, as illustrated in Figure~\ref{mae22}. 
However, the decoding process is to identify information implicitly in the premises (simulating logical deduction); thus, injecting unobserved parts is not allowed. This is the second methodological limitation, namely, an end-to-end learning process that maps the premises to the conclusion introduces contradictory training targets between the semantic encoding (for object recognition) and decoding (for logical reasoning) -- the object recognition component may introduce new parts that do not exist in the input images, and the logical reasoning component does not allow this, but can neither stop nor notice this.
For example, the Siamese networks (object recognition components) of Euler Net may inject red and blue circles into the input images that only have single green circles, and lead the reasoning component to output ``{\color{blue}{blue circle}} contains {\color{red}{red circle}}'', but the input images have only two single {\color{darkgreen}{green circles}}, as shown in Figure~\ref{br}. 

Machine learning systems with linguistic inputs face the same limitation, manifested as reasoning performance biased by surface lexical semantics.  Recent studies show that LLM reasoning can be influenced by lexical semantics, even when its underlying logical structure remains unchanged~\cite{lampinen2024language,bertolazzi-etal-2024-systematic,ozeki-etal-2024-exploring,ryb-etal-2022-analog}. Their performance is unstable for logically equivalent statements with different words or syntactic forms. In syllogistic reasoning, LLMs are more likely to accept believable conclusions and reject unbelievable ones. Chain-of-thought prompting and fine-tuning can mitigate these side effects from surface lexical semantics by decomposing long-chained reasoning into several steps, but they do not improve the reasoning performance of single-step reasoning, such as syllogistic reasoning.

\section{Experiments on The Limits of Scaling Laws}
\label{exp} 

We conducted a series of experiments to confirm that the gap between the reasoning capabilities of data-driven machine-learning systems, with either linguistic or visual inputs, and the symbolic-level logical reasoning cannot be filled by increasing the amount of training data. 



\begin{table*}[t]
\caption{Syllogistic reasoning performance of OpenAI GPT-5-nano and GPT-5.  `$\checkmark$EXPL' for correct explanation, `{\color{gray}\ding{55} H}' for hallucinative explanation. The `\#correct decision-\ding{55} H' column means a correct decision with a wrong explanation; the `\#wrong decision-$\checkmark$EXPL' column means a wrong decision with a correct explanation.}
  \label{gpt5sphnn}
  \centering
  \begin{tabular}{cclllrr}
\toprule
 & &\multicolumn{2}{c}{\#correct decision}&\multicolumn{2}{c}{\#wrong decision}&\\
 \cmidrule{3-4}\cmidrule{5-6} version& surface form      & $\checkmark$EXPL &  {\color{gray}\ding{55} H} & {\color{gray}$\checkmark$EXPL} & {\color{gray}\ding{55}H}&\#simple acc\\
\midrule 
 GPT-5-nano& words &  160 ( 62.5\%)  & {\color{gray}{90}}    &{\color{gray}{5 }}  & {\color{gray}{1}}  & 97.7\%\\ 
 & double words &  {\bf 230 ( 89.4\%)}  & {\color{gray}{22}}    &{\color{gray}{4}}  & {\color{gray}{0}}  & 98.4\%\\
& simple symbols &  226 ( 88.3\%)  & {\color{gray}{24}}   
 &{\color{gray}{6}}  & {\color{gray}{0}}   & 97.7\%\\
& long random symbols &  222 ( 86.7\%)  & {\color{gray}{25}}   
 &{\color{gray}{9}}  & {\color{gray}{0}}  & 96.5\% \\
    \hline
GPT-5& words &  {\bf 239 ( 93.4\%)}  & {\color{gray}{16}}   
  &{\color{gray}{1}}  & {\color{gray}{0}}  &99.6\%\\ 
 & double words &  234 ( 91.4\%)  & {\color{gray}{21}}   &{\color{gray}{1}}  & {\color{gray}{0}}  & 99.6\%\\
& simple symbols &  236 ( 92.2\%)  & {\color{gray}{15}}   
  & {\color{gray}{5}}  & {\color{gray}{0}}  &98.0\% \\
& long random symbols &  231 ( 90.2\%)  & {\color{gray}{25}}   
&{\color{gray}{0}}  & {\color{gray}{0}}  & {\bf 100.0\%} \\
\bottomrule
  \end{tabular}
\end{table*} 

\subsection{Experiment 1}

\paragraph{The aim} We investigate whether data-driven machine-learning systems with linguistic inputs can achieve symbolic-level logical reasoning, focusing on two recent and readily accessible OpenAI models: GPT-5 and GPT-5-nano. 

\paragraph{Setting of the experiment}  We follow the method in \citep{djl2025} that used syllogistic statements with four surface lexical patterns: (1) meaningful words, e.g. {\em Socrates}, (2) doublele words, e.g. {\em City\_Socrates}, (3) simple symbols, e.g. {\em S}, and (4) long random symbols, e.g. {\em VnWKvqcBsEh1}, to determine the satisfiability of all 256 types of classic Aristotelian syllogistic reasoning. The motivation for introducing the pattern of double words is to enhance the probability of lexical semantics to affect reasoning.

\paragraph{Evaluation metrics} We use two evaluation metrics: (1) the normal metrics in terms of accuracy (the \#simple acc column in Table~\ref{gpt5sphnn}); (2) the metrics of reaching symbolic-level logical reasoning, namely, a response is correct if both the decision and the explanation are correct (the \#correct decision-$\checkmark$EXPL column in Table~\ref{gpt5sphnn}). 

\paragraph{Results} The experiment's results confirm the high performance of both OpenAI's GPT-5-nano and GPT-5 in syllogistic reasoning. Meanwhile, experiments show that surface lexes affect reasoning performance, and each surface pattern made at least 15 correct decisions with incorrect explanations. When fed double-word statements, GPT-5-nano achieved 89.4\% correct decisions with correct explanations, better than other forms. With single-word statements, GPT-5 achieved 93.4\% correct decisions with correct explanations, better than using other forms. In particular, GPT-5 achieved 100\% correct decisions with long random symbols, but 25 correct decisions were supported by wrong explanations. 
\paragraph{Conclusion} Because 100\% accuracy represents the performance ceiling and terminates the optimisation process, attaining symbolic-level performance remains impossible.

\begin{figure*}[h!]  
  \centering  
\includegraphics[width=1\textwidth]{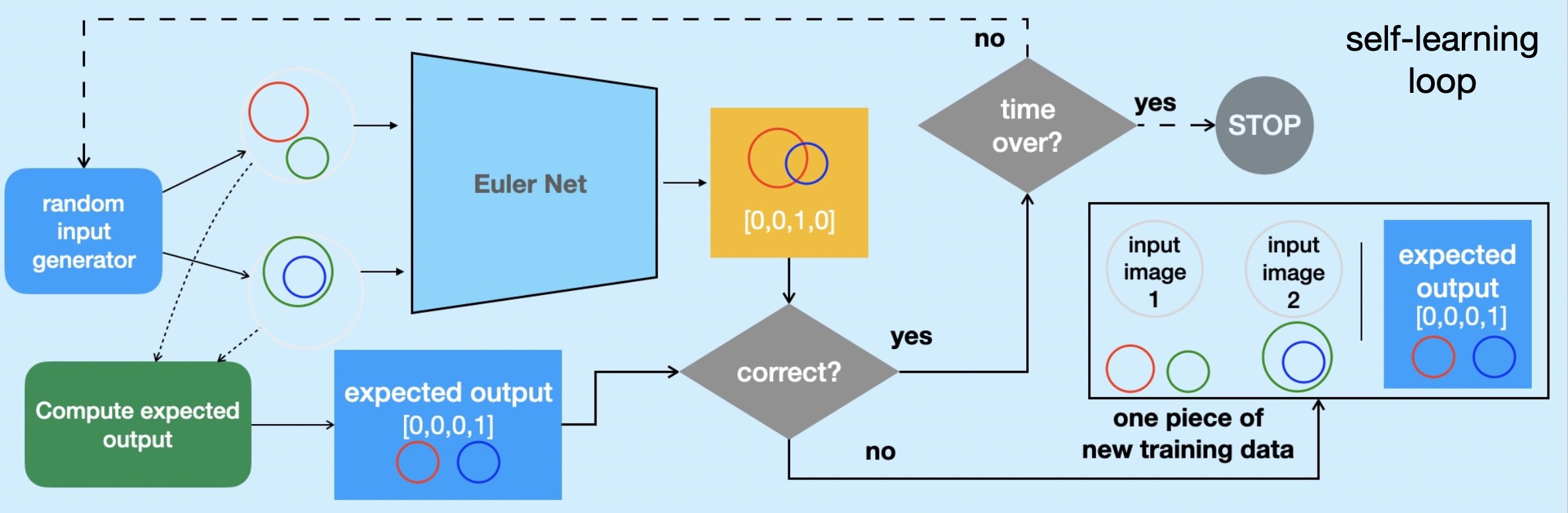}
  \caption{The architecture of Super Euler Net.}
  \label{ext_en}
\end{figure*}

\subsection{Experiment 2}

\paragraph{The aim} We implement a self-learning mechanism for Euler Net, which can automatically create training data for errors the systems make. We examine whether EulerNet, equipped with this mechanism, can continuously improve its performance and converge arbitrarily closely to symbolic-level logical reasoning. 
 
\paragraph{The self-learning mechanism} 

In the setting of Euler Net, all image inputs are circles; therefore, the correct outputs of syllogistic reasoning can be computed. We introduce a light-weighted self-learning mechanism that allows Euler Net to automatically identify incorrect output and create new training data accordingly, as follows: (1) the self-learning system randomly generates image-input for Euler Net and compute the correct reasoning result; (2) it feeds the input to Euler Net and checks whether Euler Net's output is correct; (3) If not correct, a new piece of input-output pair in Step (1) will be a new piece of training data. A self-learning loop consists of two stages: identify enough errors and create new training data; then, train the Euler Net with this new training data. 

Specifically, in the random generation step, the central point and the radius of a circle are random, with two restrictions as follows: (1) circles are fully inside the boundary of an image; (2) the minimum radius is 0.1. We allow all possible combinations between two circles. Based on these criteria, Euler Net creates a new test data set $\mathcal{D}_1$.
\begin{figure}[h] 
  \centering  
\includegraphics[width=1\textwidth]{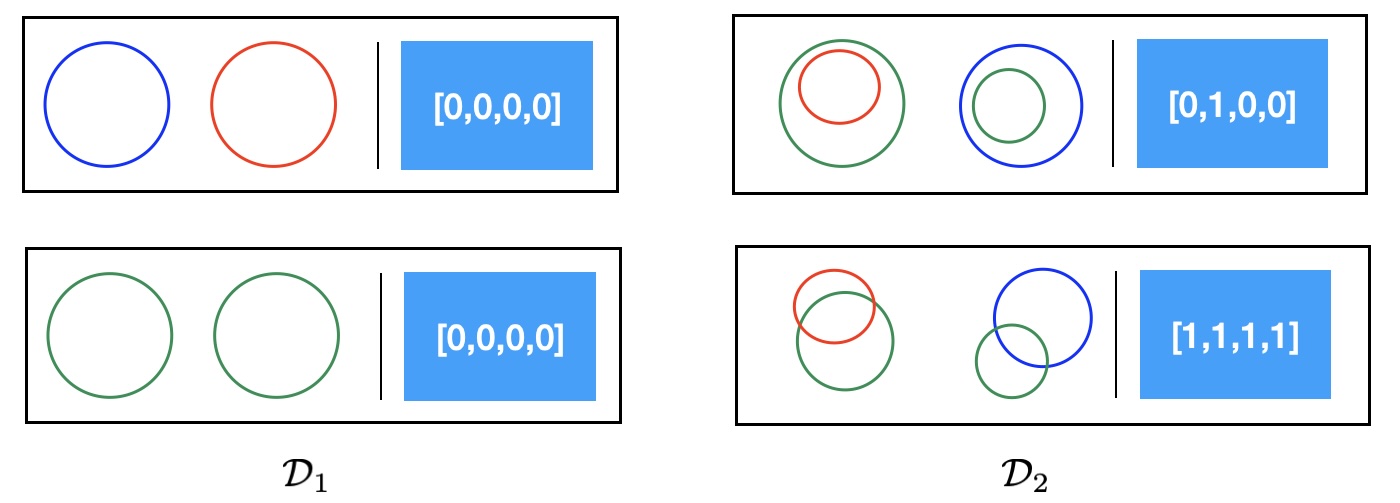}
  \caption{$\mathcal{D}_1$ is a newly created dataset; $\mathcal{D}_2$ is randomly selected from the standard training dataset.}
  \label{training1}
\end{figure}
\begin{figure}[h!]  
\centering
\includegraphics[width=1\textwidth]{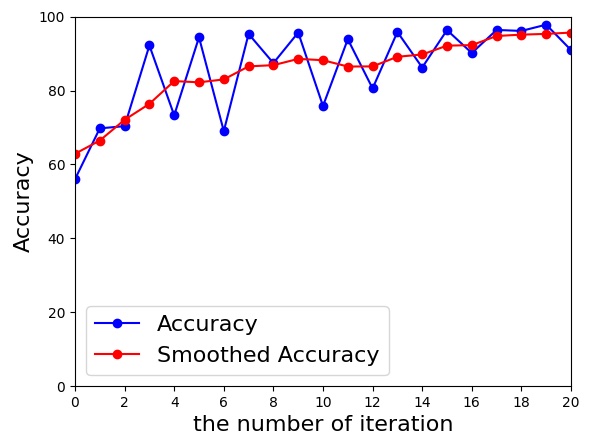}
\caption{The accuracy of SupEN at each iteration with newly generated training data.} 
\label{fig_accuracy} 
\end{figure} 

In the training stage, Euler Net creates a training data $\mathcal{D}_T$ that consists of newly created training data $\mathcal{D}_1$ and original training data $\mathcal{D}_2$,    $\mathcal{D}_T=\mathcal{D}_1\cup\mathcal{D}_2$, as shown in Figure~\ref{training1}. For example, $\mathcal{D}_1$ is a newly created dataset with one-circle images. The size of $\mathcal{D}_2$ is 9 times larger than that of $\mathcal{D}_1$. 
\paragraph{Setting of the experiment} Different from standard statistical learning, where testing data and training data have the same or similar distribution~\citep{Bengio22}, in this experiment, the testing data are randomly generated. The experiment was started by using a well-trained Euler Net~\cite{WangJL18}. Being fed with randomly generated testing data, it reached the accuracy of $56.0\%$.  

\paragraph{Experiment result} 
We let Euler Net loop 20 times through the searching-training process to improve its reasoning performance. Through repeated training on newly created datasets, accuracy improves steadily and reaches a peak of $97.8\%$ after the 19th iteration, as illustrated in Figure~\ref{fig_accuracy}.  The increase in accuracy supports the law of scaling, but with the gap of $100.0\%-97.8\% = 2.2\%$, Euler Net has not achieved the symbolic-level performance. 

\subsection{Experiment 3}
\paragraph{The aim} The training data in Experiment 2 were generated by the combination table, which cannot distinguish each valid type of syllogistic reasoning. This experiment evaluates the final Euler Net from Experiment 2 as a reasoning engine for syllogistic tasks using linguistic inputs designed to distinguish each valid syllogistic reasoning.

\paragraph{Setting of the experiment}  We created a new testing dataset as follows: For each valid type, we created 500 different premises by extracting hypernym relations from WordNet-3.0 \citep{MillerWordNet95}. For each premise, we deduce its valid logical conclusion and its negation.
In the hypernym structure,  {\em elementary\_particle.n.01} is a descendent of {\em  natural\_object.n.01} and {\em artifact.n.01} is not a descendent of {\em  natural\_object.n.01}, we create the valid syllogistic reasoning and its negation, as follows. 
\syllogism{{\em all elementary\_particle.n.01 are natural\_object.n.01.}\\{\em no artifact.n.01 are natural\_object.n.01.}}
{{\em no elementary\_particle.n.01 are artifact.n.01.}}
\syllogism{{\em all elementary\_particle.n.01 are natural\_object.n.01.}\\ {\em some artifact.n.01 are natural\_object.n.01.}}
{{\em some elementary\_particle.n.01 are artifact.n.01.}} 
We group equivalent syllogistic statements together; for example, {\em no x are y} and {\em no y are x} are in the same group. This reduces 24 {\em valid} syllogism types into 14 groups, totalling 14000 syllogism reasoning tasks. We use the pre-processing tool of the original Euler Net to transform premises into coloured circles and conclusions into vectors.  
 
\paragraph{Experiment result} We fed the new dataset to the Euler Net with 19 loops of self-learning. It achieves 100\% accuracy across 8 valid types of syllogistic reasoning. Accuracies of the remaining 16 types range from $50\%$ to $83.3\%$. The overall accuracy is 76\%, as shown in the Appendix. This performance is consistent with~\cite{Eisape2024}'s evaluation with PaLM 2 and Llama 2 family LLMs --- the best performance was achieved by PaLM 2-Small, with an accuracy of about 75\%. These results suggest that reasoning from image inputs is comparatively coarse-grained, with performance on individual syllogism types falling below that of machine learning systems using linguistic inputs.

\begin{figure*}[h] 
  \centering  
\includegraphics[width=1\textwidth]{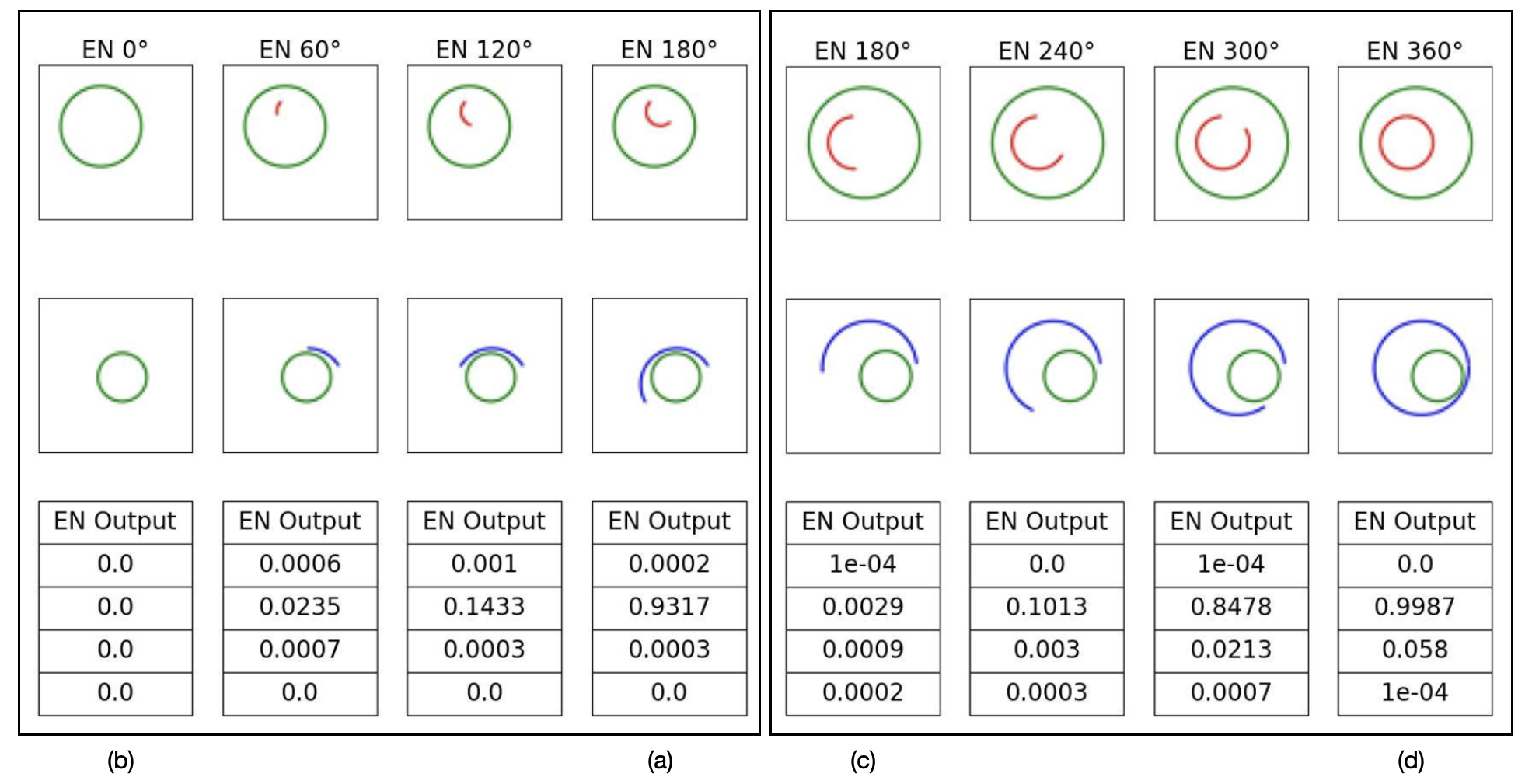}
  \caption{(a) Euler Net may automatically complete a semicircle into a full circle (with the output $[0.0002, 0.9317, 0.0003, 0]$). As we decrease the arc length to $120^\circ$, $60^\circ$, and $0^\circ$, the value decreases accordingly. (b) Euler Net may automatically ignore the half circle and only take one green circle as input (with the output $[0.0001, 0.0029, 0.0009, 0.0002]$). As we increase the length of the arc to $240^\circ$, $300^\circ$, and $360^\circ$, the corresponding values increase accordingly.} 
  \label{ex1}
\end{figure*} 
\subsection{Experiment 4}

\paragraph{The aim} Though reasoning from image inputs is comparatively coarse-grained, there were still 8 types for which Euler Net achieved 100\% accuracy. We show that Euler Net cannot achieve symbolic-level reasoning for one of these 8 types, due to the second methodological limitation. The remedy is that when the encoding component injects new parts, Euler Net shall flag the input as \emph{unintended-input}. We train EulerNet to classify single-circle inputs as members of the \emph{unintended-input} class and show that this class cannot be exhaustively enumerated.
 
\paragraph{Setting of the experiment} We define a new output vector $[0,0,0,0]$ representing unintended inputs and train EulerNet to classify single-circled inputs into $[0,0,0,0]$ till it reaches 100\% accuracy. In the end, 
EulerNet can perform syllogistic reasoning for regular inputs and classify single-circle inputs as unintended. 
Then, we introduce a new unintended input: one image contains a green circle and a red semicircle, while the other contains a blue semicircle and a green circle.

\paragraph{Experiment results} 
Experiments show that sometimes EulerNet completes the two half-circles into two whole circles, and concludes $[0, 1, 0, 0]$ {\em the blue circle contains the red circle}, as shown in Figure~\ref{ex1}(a). In this case, if we decrease the arc length of the two half-circles, the confidence value flagging {\em the blue circle contains the red circle} will decrease, as shown in the input images and outputs from Figure~\ref{ex1}(a) to (b). Sometimes, EulerNet simplified one green circle and a half circle into one green circle (half circles are neglected) and concludes the inputs are unintended $[0,0,0,0]$, as shown in Figure~\ref{ex1}(c). In this case, if we increase the arc length of the two half-circles, the confidence value flagging {\em the blue circle contains the red circle} will increase correspondingly, as shown in input images and outputs from Figure~\ref{ex1}(c) to (d). 
This, however, will automatically create another kind of unintended pattern: one green circle and a partial circle with $(180^\circ+360^\circ)/2=270^\circ$. This loop will never end.  Training data cannot exhaust all unintended inputs, because new training data generate new unintended inputs. Therefore, even if Euler Net achieves 100\% accuracy after sufficient self-learning iterations, unintended inputs that it cannot reason for sure will inevitably remain.
\section{Conclusion}
\label{gap}
 
Recent studies show that, despite impressive empirical performance, LLMs have not attained the symbolic-level syllogistic reasoning. Despite the simplicity of syllogistic reasoning, two methodological limitations prevent data-driven systems from achieving this level. Our analysis focuses on training data—the fuel of data-driven machine learning systems—and concludes that no desired fuel can provide sufficient information for them to reach the symbolic level of syllogistic reasoning, even for a single valid type. Linguistic input machine learning systems suffer from lexical semantics, while image input systems suffer from unintended inputs. Since syllogistic reasoning underpins logical reasoning and human rationality, this limitation has broader implications for the reasoning capabilities of data-driven machine learning systems and enhances our understanding of the scaling law.

\section*{Acknowledgement}

We gratefully acknowledge Björn Gintzel for his coding and implementation support.


  

\bibstyle{sn-mathphys-num}
\bibliography{XBib,XBib_NN_s}

 
\appendix

\newpage
\section{The list of 24 valid types of syllogistic reasoning}
\label{vtype}
\begin{table}[h]
 \caption{List of all 24 valid syllogisms, each having a name whose vowels indicate types of moods, e.g., vowels in `C\underline{E}L\underline{A}R\underline{E}NT' indicate {\em universal negative} (E), {\em universal affirmative} (A), and {\em universal negative} (E).} 
\label{24types}
\centering
  \begin{tabular}{clll}
    \hline 
   Num & Name     & Premise &Conclusion     \\
    \hline 
   1& BARBARA & all $X$ are $Y$, all $Y$ are $Z$ & all $X$ are $Z$    \\
   2& BARBARI & all $X$ are $Y$, all $Y$ are $Z$ & some $X$ are $Z$     \\
   3&CELARENT& no $Y$ is $Z$, all $X$ are $Y$ & no $X$ is $Z$     \\
    4& CESARE& no $Z$ is $Y$, all $X$ are $Y$ & no $X$ is $Z$      \\
   5& CALEMES &  all $Z$ are $Y$, no $Y$ is $X$ & no $X$ is $Z$ \\
   6& CAMESTRES & all $Z$ are $Y$, no $X$ is $Y$ & no $X$ is $Z$  \\
   7& DARII & all $Y$ are $Z$, some $X$ are $Y$ & some $X$ are $Z$    \\
   8&DATISI& all $Y$ are $Z$, some $Y$ are $X$ & some $X$ are $Z$     \\
   9&  DARAPTI & all $Y$ are $X$, all $Y$ are $Z$ & some $X$ are $Z$       \\
   10& DISAMIS & some $Y$ are $Z$, all $Y$ are $X$ & some $X$ are $Z$       \\
  11& DIMATIS &  some $Z$ are $Y$, all $Y$ are $X$ & some $X$ are $Z$      \\
   12& BAROCO& all $Z$ is $Y$, some $X$ are not $Y$ & some $X$ are not $Z$     \\
    13& CESARO& no $Z$ is $Y$, all $X$ are $Y$ & some $X$ are not $Z$      \\
   14&CAMESTROS & all $X$ are $Y$, no $Y$ is $Z$ & some $X$ are not $Z$      \\
   15&CELARONT& no $X$ is $Y$, all $Z$ are $Y$ & some $X$ are not $Z$     \\
   16& CALEMOS& all $Z$ are $Y$, no $Y$ is $X$ & some $X$ are not $Z$     \\
   17& BOCARDO & some $Y$ are not $Z$, all $Y$ are $X$ & some $X$ are not $Z$       \\
   18& BAMALIP & all $Y$ are $X$, all $Z$ are $Y$ & some $X$ are $Z$       \\
   19&FERIO& some $X$ are $Y$, no $Y$ is $Z$ & some $X$ are not $Z$     \\
   20&FESTINO& some $X$ are $Y$, no $Z$ is $Y$ & some $X$ are not $Z$      \\
   21& FERISON& some $Y$ are $X$, no $Y$ is $Z$ & some $X$ are not $Z$      \\
  22&FRESISON& some $Y$ are $X$, no $Z$ is $Y$ & some $X$ are not $Z$     \\

   23&FELAPTON& all $Y$ are $X$, no $Y$ is $Z$ & some $X$ are not $Z$      \\
   24&FESAPO& all $Y$ are $X$, no $Z$ is $Y$ & some $X$ are not $Z$     \\
 \hline 
  \end{tabular} 
\end{table}

\end{document}